# 3D Point Cloud from Close-Range Photogrammetry for Defect Characterisation of Rubberised Concrete

Jiacheng Liu[1], Mohammed Alnahhal[1], Ailar Hajimohammadi[1], Sara Gonizzi Barsanti[2], Jinling Wang[1], Mohsen Kalantari[1, *]

1: School of Civil and Environmental Engineering, University of New South Wales, Sydney, NSW 2052, Australia – Jiacheng.liu4@unsw.edu.au; m.alnahhal@unsw.edu.au; ailar.hm@unsw.edu.au; jinling.wang@unsw.edu.au; mohsen.kalantari@unsw.edu.au

2: Dept. of Engineering, Università degli Studi della Campania Luigi Vanvitelli, via Roma 29, 81031, Aversa (CE), Italy - sara.gonizzibarsanti@unicampania.it

**Abstract**

While three-dimensional (3D) point clouds are widely used in civil engineering, mainstream LiDAR systems such as Terrestrial Laser Scanning (TLS) are physically constrained to laboratory environments. Since their laser spot size typically exceeds the width of microcracks, the beam physically bridges over voids, rendering TLS unsuitable for fine-scale defect analysis. Alternatively, close-range photogrammetry utilising Structure-from-Motion (SfM) and Multi-View Stereo (MVS) algorithms offers a solution for testing highly tortuous materials, and its utility at fine-scale remains underexplored. This study adapts photogrammetric workflows specifically for rubberised concrete (RuC), a sustainable composite exhibiting high ductility and complex fracture morphologies. High-resolution image sets were captured using a Canon DSLR and an iPhone 16 to generate dense 3D models. Comparisons revealed that the DSLR-based reconstruction achieved sub-millimetre resolution, demonstrating superior performance for fine-scale surface monitoring. An RGB-guided crack extraction method was developed to enhance the identification of surface defects and isolate potential crack areas from the background. The extracted crack regions were visually distinguishable and provided a well-structured geometrical representation of defect morphology. Furthermore, a Pre and Post-Test deformation analysis was conducted to quantify surface displacement across testing stages. The results confirm that this close-range photogrammetry workflow is a flexible, high-resolution alternative to LiDAR for surface inspection and deformation monitoring of specimens in laboratory settings. Ultimately, this approach establishes a robust geometric baseline for future automated 3D feature characterisation and material performance evaluation.

## 1. Introduction

The global consumption of rubber reached nearly 24.9 million tons in 2010 and continues to increase (Gerges et al., 2018) while half of the rubber is improperly managed (Liang et al., 2025). The challenge of disposing of waste from end-of-life tyres is driving research into sustainable alternatives to conventional construction materials. One promising solution is rubberised concrete (RuC), also well known as crumbed rubber concrete (CRC), which utilises recycled tyre rubber as a partial replacement for traditional concrete aggregates. This utilisation not only provides a value-added application for waste but improves the properties of concrete. For example, RuC exhibits enhanced ductility, impact resistance, and energy absorption capacity, but often compromises strength capacity compared to conventional concrete (Elshazly et al., 2020). At the same time, the optimal mix design to balance rubber content, particle size, and admixtures is still under investigation. Furthermore, understanding the fracture mechanics of cementitious composites is an important path for developing more reliable construction mixture designs.

A key indicator of RuC performance under flexural loading is the generation and behaviour of cracks. Accurate identification and quantification of cracks in specimens produced by bending tests are critical for validating their performance and refining material models. However, quantifying these minor features, which are often at millimetre level or less, is challenging. Traditional post-test analysis, which primarily relies on visual inspection or contact measurement using measuring tapes, straightedges, and microscopes, is labour-intensive and fails to capture the complete 3D topology of the fracture surface (Ren et al., 2025; Xu et al., 2020). With the rapid development of hardware and software in Computer Vision (CV), several studies have been conducted to improve the automation and efficiency of civil engineering inspections: digital image processing, 3D LiDAR scanning technology (i.e., point cloud data), and Structure from Motion (SfM) are widely used in the building industry (Ren et al., 2025). Point cloud, in particular, offers significant flexibility and generalisation for macroscopic structure classification and overall health monitoring (Loverdos and Sarhosis, 2024).

While existing macro-scale methods are highly effective for large structures, their spatial resolution severely limits the extraction of fine detail. For instance, an automated crack detection method using median filters and locally adaptive thresholding algorithms to map point cloud collected from Leica C10 TLS achieved an accuracy of only 10-38 $mm$ when compared against total station surveys (Rabah et al., 2013). Similarly, Błaszczak-Bąk et al. (2023) highlight the inherent limitations of physical hardware, noting that the laser spot size of a TLS is typically 3-5 mm. While this error margin is perfectly acceptable for assessing massive structural damage in the field, it exposes a significant technological gap in fine-scale fracture analysis.

In practice, large-scale and significant defects do not always appear; therefore, laboratory testing remains another essential branch of civil engineering practice that requires modelling and analysing object defects at the sub-millimetre level. The data and behaviours acquired in the lab can also serve as a benchmark for assessing the quality of future real-world construction projects. A critical aspect of lab testing is surface defect characterisation during flexural testing, which demands precise measurement of imperfection down to sub-millimetre level. However, as mentioned before, commercially available TLS and MLS LiDAR scanning systems typically have a point resolution of up to 1 $mm$ or greater (Gabara and Sawicki, 2023; Suchocki et al., 2025), making them unsuitable for detailed analysis of fine-scale fracture surfaces and incipient cracks, where millimetre-level accuracy is required. Handheld Laser Scanners (HLS) and high-resolution handheld digital microscopes (HDM) are considered good candidates for laboratory testing (Suchocki et al., 2025). Tests with these devices identified that a 0.1 $mm$ resolution in the scanning process is sufficient to identify defects at the sub-millimetre scale. The primary limitations of these systems, apart from the prohibitive hardware cost, are the strict requirements to place coded targets throughout the object to enable spatial position calculation and low resolution of RGB information, which is challenging for the segmentation process. While 3D point clouds are gaining traction for assessing conventional rigid concrete, their application in material laboratory testing remains infrequent. Common approaches for concrete in the lab often rely on 2D digital image correlation (DIC) or spatial algorithms optimised for relatively planar and brittle fracture surfaces. However, RuC presents unique topological challenges. The inclusion of waste tyre rubber introduces high ductility and elastic inclusion pullout, resulting in highly tortuous and irregular 3D fracture morphologies. Consequently, the traditional 2D imaging method inherently fails to capture the volumetric severity of these defects, necessitating a transition to high-resolution 3D point cloud analysis. Due to this geometric complexity, the specific algorithmic framework required to map and quantify high-resolution point cloud data for RuC accurately has not yet been fully explored. Therefore, a primary motivation of this study is to address this distinct gap by establishing a clear, highly accessible, and systematic workflow tailored specifically to the complex fracture surfaces of rubber composite materials.

To address these gaps, this study utilises close-range photogrammetry to generate high-resolution 3D point clouds of RuC specimens after the bending test. By capturing a series of high-resolution, overlapping digital images from multiple perspectives, SfM and Multi-View Stereo (MVS) algorithms can reconstruct a metrically accurate 3D model of the specimen surface. This paper details the workflow and use of this technique and methodology, implemented in Agisoft Metashape 3D reconstruction software, to generate dense point clouds with sub-millimetre resolution, making it a viable alternative for the precise geometric characterisation of cracks on RuC surfaces. This software has been chosen for its reliability and the semi-automatic approach that allows the chosen settings to be improved during processing (Feddema and Chiu, 2024; Kostrzewa et al., 2025; Trombini et al., 2025).

The experimental investigations in this study are based on laboratory mechanical testing. The loading process generates several types of cracks, which serve as indicators of the mechanical properties of the corresponding mixing design. The primary contributions of this paper can be summarised as follows:

1. The feasibility of the 3D point clouds from close-range photogrammetry for characterising the minor surface defects is validated, establishing a foundation for batch 3D feature quantification.
2. A systematic workflow of high-resolution 3D reconstruction data acquisition and processing for small-scale specimens in laboratory applications is proposed to ensure metric accuracy at the sub-millimetre level.
3. Demonstrate the utility of the generated point cloud data for both quantitative analysis (e.g., crack length estimation, deviation analysis) and enhanced visual communication. The complete dataset collected during this study will be made publicly available to support future research.

## 2. Related work

Recently, 3D point cloud data, mostly acquired from TLS and MLS, have become fundamental tools and data sources for structural health monitoring and quality assessment in civil engineering (Blaskow and Maas, 2024; Chang et al., 2024; Wang et al., 2024; Zhu et al., 2011). These implementations cover infrastructure monitoring, deformation between as planned and as-built, defect feature extraction and quantification, and building information modelling. The primary advantage of these applications lies in their ability to efficiently capture 3D spatial geometry data, offering a comprehensive and intuitive approach compared to traditional inspection methods.

### 2.1 Point Cloud Applications in Civil Engineering

For defect analysis, an automatic quality control framework for prefabricated concrete components using K-Nearest Neighbour (KNN) classification and Delaunay triangulation was developed to identify and measure surface defects from point cloud data. This approach demonstrated the feasibility of automated, geometry-driven defect assessment (Xu et al., 2020). Recently, a point cloud-based quality inspection system for precast concrete components was introduced. This system is able to assess geometric irregularities, positional accuracy, surface flatness, and crack presence (Ren et al., 2025). This trend extends to complex inspection tasks, Wang et al., (2024) utilised the Random Sample Consensus (RANSAC) algorithm to decompose reinforced concrete slab point clouds into distinct structural elements, including the formwork, upper and lower reinforcement lattices, and individual rebars. The method achieved a concrete cover thickness measurement accuracy of 98.4%, underscoring the high precision attainable through geometric segmentation of point clouds. A volumetric defect quantification method based on point cloud slicing was proposed by Xu et al., (2025). The workflow included defect area extraction using local curvature and point-to-plane distance calculations, followed by Euclidean clustering to classify the slices. This method proved the effectiveness of 3D models for large-scale infrastructure defect detection and volume estimation.

Nowadays, Deep Learning (DL) and CV techniques have become essential tools for analysing point cloud data. Chang et al., (2024). developed an automated rebar inspection framework that integrates DL, CV, and Building Information Modelling (BIM). The method combines Mask R-CNN segmentation with a clustering algorithm for spatial classification, achieving over 90% accuracy and more than 97% recall for both main and tie rebars during 3D instance segmentation. Bolourian et al., (2023) released a benchmark point cloud dataset for detecting surface defects on concrete bridges and proposed a deep learning-based method, Surface Normal PointNet++, for this task. This approach identifies normal vector orientations and defect depths, thereby enhancing automated defect detection compared to traditional methods. These studies demonstrate the promising effectiveness and progress of using advanced algorithms to extract and analyse detailed 3D geometrical information from large-scale point cloud data for quality and structural assessments.

### 2.2 The limitation of lidar for fine-scale defect analysis

3D point clouds acquired from LiDAR-based systems have been successfully deployed for macro-scale structural health monitoring and defect assessment across massive infrastructure. However, 3D data acquisition via conventional LiDAR systems, such as TLS and MLS, presents a physical bottleneck when transitioning LiDAR-based methods to fine-scale defect analysis in laboratory settings.
As highlighted by Błaszczak-Bąk et al., (2023) and Laefer et al., (2014), the inherent laser spot size of TLS typically ranges from $3mm$ to $5mm$, strictly limiting the minimum detectable crack width. Consequently, when applied to fine-scale concrete specimens in the lab, the laser beam diameter physically bridges over sub-millimetre microcracks. For example, recent laboratory studies evaluating TLS during concrete beam flexural testing, Janowski et al., (2016) demonstrated that while global structural deflection is captured accurately, the point cloud density and spatial footprint inherently smooth out fine-scale incipient cracks. Even when applying advanced extraction algorithms to TLS data, the point spacing remains a strict barrier. Rabah et al., (2013) combined 2D imaging and 3D TLS data to extract cracks, but by employing median filtering and locally adaptive thresholding, the mapping accuracy was only 10-38 $mm$. To overcome this, Suchocki et al., (2025) proposed a small crack measurement method by integrating a handheld scanner and a high-resolution digital microscope as complementary data for TLS measurement. While this hybrid method can achieve sub-millimetre measurement, it comes at the cost of operational efficiency and requires highly specific, prohibitive hardware setups. Consequently, although macroscopic TLS error margins are perfectly acceptable for assessing massive structural damage in the field, they expose a significant technological gap in sub-millimetre fracture analysis in materials science under a laboratory environment.

### 2.3 Close-range photogrammetry

To overcome the point spacing limitations mentioned above, close-range photogrammetry offers a powerful and accessible alternative for generating high-resolution 3D models with more accurate RGB information. By leveraging SfM and MVS techniques, dense, high-fidelity point clouds with sub-millimetre-level detail can be generated from overlapping 2D images. This capability makes it well-suited to capturing the subtle geometric and textural variations of surfaces.

Paul et al., (2025) proposed an improved pipeline by modifying the minimum triangulation angle, reducing the overall re-projection error, and using a tilling buffer for 3D model reconstruction. This study also compared the performance of five reconstruction software programs. It is reported that open-source software is more flexible, allowing for the addition of custom features. In contrast, commercial platforms such as Agisoft, Pix4DMapper, and Context Capture deliver high-quality, industry-standard results but offer less flexibility. The image entropy and BRISQUE Score were used to evaluate the quality of the generated 3D models, but geometric evaluation remains lacking. Gabara and Sawicki, (2023) systematically examined the challenges and error sources associated with mainstream 3D reconstruction software in close-range applications. They published the CRBeDaSet dataset, which provides a benchmark for evaluating photogrammetric reconstruction performance. The comparative analysis between TLS and photogrammetric methods highlighted the latter’s potential for high-resolution digital modelling, demonstrating comparable accuracy when proper imaging geometry and calibration are applied.

Despite these advantages, further systematic validation of the photogrammetric workflow, application, and performance under laboratory environments remains limited. In particular, the integration of 3D models with material behaviour analysis, like RuC with varying mixing designs, has yet to be systematically explored. The heterogeneous composition and distinctive cracking behaviour of RuC present both a challenge and an opportunity for validating high-resolution, image-based 3D reconstruction workflows.

Consequently, there is a specific research gap in developing a validated, systematic, and efficient photogrammetric workflow for characterising minor surface defects. Furthermore, the capacity of fine-scale defect quantification using a 3D model to link defect behaviour with the corresponding mix design remains largely unexplored. This study seeks to address these gaps by investigating the performance, applicability, and accuracy of close-range photogrammetric reconstruction for surface defect analysis in RuC specimens.

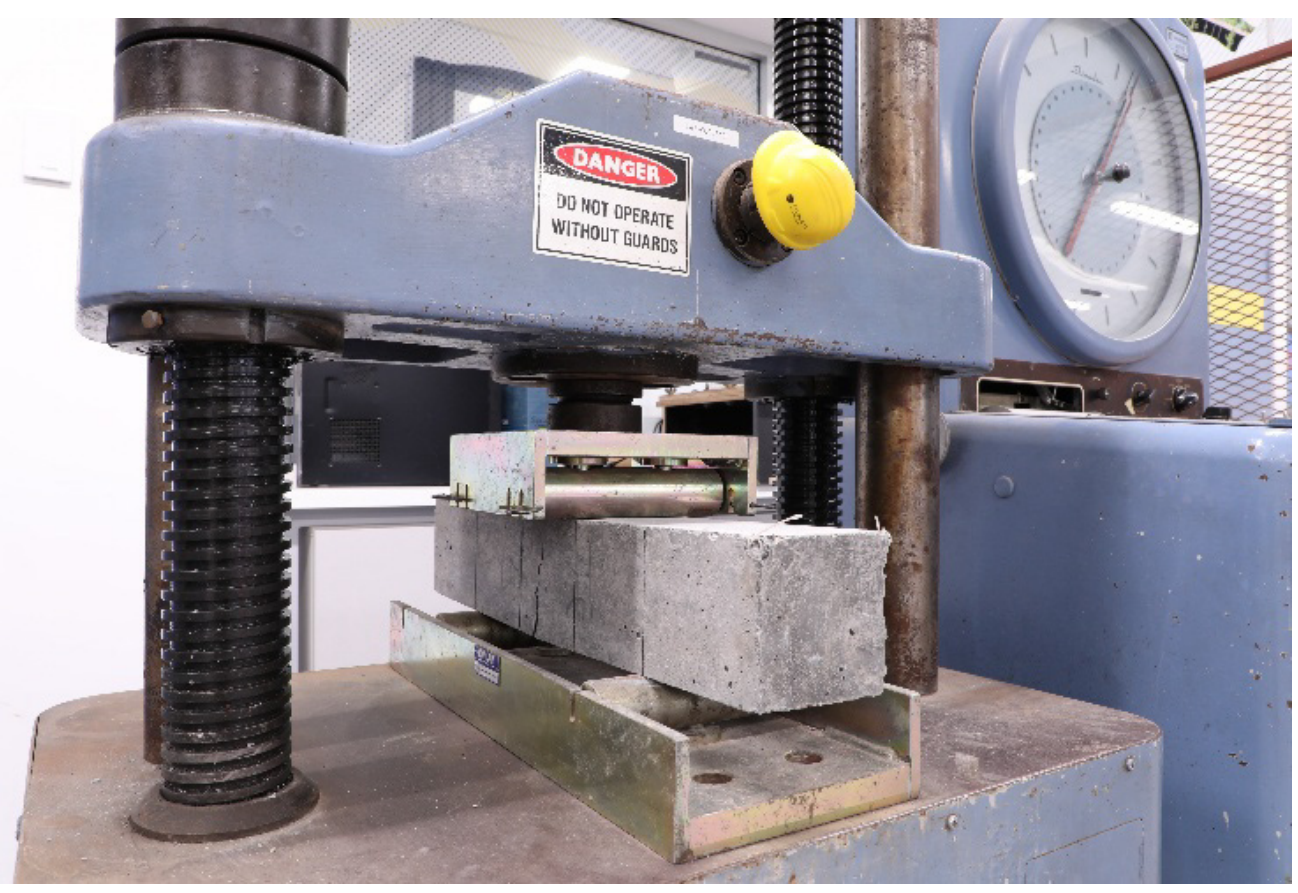

Figure 1 Bending test setting

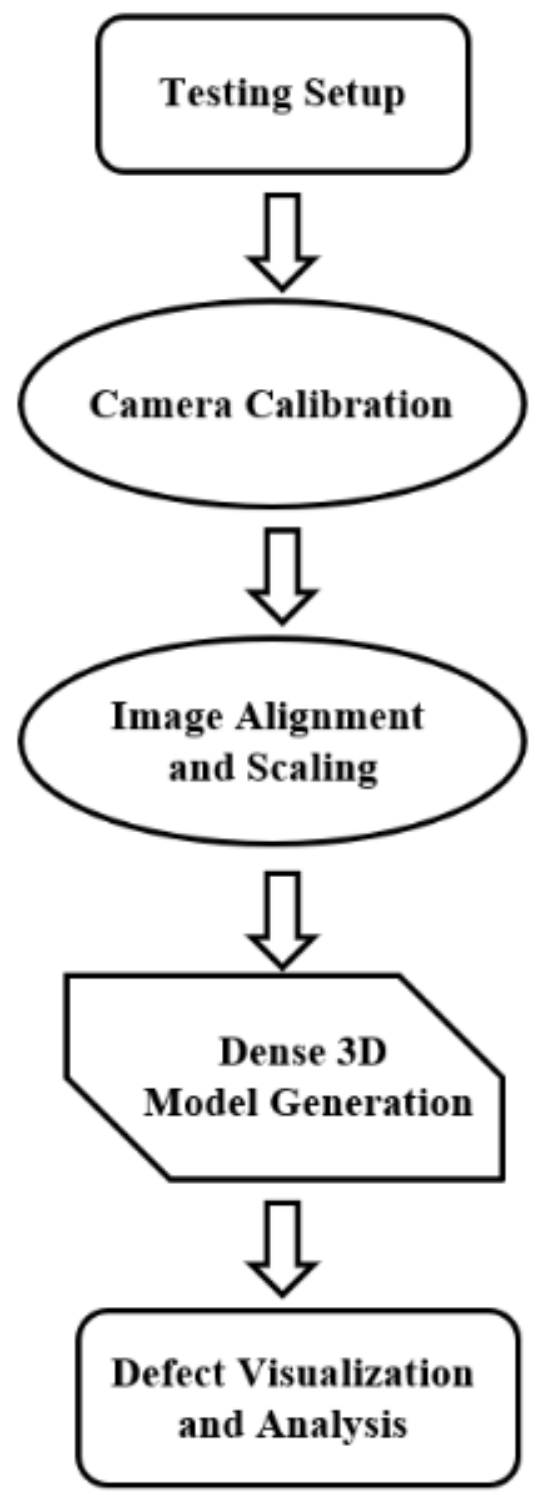

Figure 2 Workflow illustration

## 3. Methodology

### 3.1 Specimen preparation and bending test

In this study, RuC prism beams measuring 550 $mm$ × 100 $mm$ × 100 $mm$ were prepared as test samples. The concrete mix consisted of fine aggregate (river sand), coarse aggregate (10 $mm$ basalt sand), and 5 $mm$ shredded rubber sourced from waste tyres. The beams were tested to failure under four-point flexural bending (in accordance with Australian Standard AS 1012.11) using a SHIMADZU universal testing machine. The experimental setup is illustrated in **Figure 1**.

### 3.2 Data collection

The results of the final 3D reconstruction depend heavily on the quality of the input images. Therefore, a solid and systematic image acquisition protocol is essential to ensure the geometric accuracy and metric reliability of the reconstructed model. The following procedure (also shown in **Figure 2**) describes the image acquisition process used to capture the surfaces of cracked RuC beam specimens in this study.

To evaluate the reconstruction quality across different image sources, a Canon EOS 5D Mark IV DSLR and an iPhone 16 were used to capture images (**Figure 3**). The DSLR features a 30.4-megapixel sensor measuring 36.0 $mm$ × 24.0 $mm$ with a physical pixel size of 5.36 $\mu m$. For this camera, a 24 $mm$ prime lens was selected to ensure consistent geometric characteristics across the imaging process. Conversely, the iPhone 16 is equipped with a 48-megapixel sensor, a 1.0 $um$ physical pixel size and a fixed $f/1.6$ aperture with a 26 $mm$ equivalent focal length. Since the iPhone cannot manually adjust its focal length or aperture, the images were captured using the default camera settings. Based on the setup above, the Ground Sample Distance (GSD) of the Canon camera is about 0.15 mm. GSD can be calculated by the following formula:

$$GSD = \frac{p \cdot H}{f}$$

Where $p$ is the physical single pixel on the camera sensor, $H$ is the spatial distance between the camera optical centre and the object being measured (700 $mm$ in this case), and $f$ is the focal length of the lens. Although modern SfM algorithms can estimate camera parameters during bundle adjustment, the intrinsic parameters, including focal length, principal point, and lens distortion coefficients, were pre-calibrated to achieve higher metric fidelity.

**Camera settings**: To ensure consistency and repeatability across all images, the camera was set to manual mode. The ISO was fixed at 250 to minimise noise, and the aperture was set to $f/10$ to provide a large depth of field, and the shutter speed was adjusted as needed to maintain proper exposure. Images were recorded in both RAW and JPG formats to preserve the full radiometric range and enable flexible post-processing adjustments.

**Lighting conditions**: A key challenge in imaging samples is the presence of specular surfaces. To achieve shadowless, diffuse illumination, an external flash with a shadow board was positioned on the opposite side of the specimen. This setup reduces harsh shadows that could be misinterpreted as cracks by the follow-up processing.

**Image capture strategy**: An orbital acquisition pattern (illustrated in **Figure 4**) was adopted, in which the camera was moved around the specimen at multiple viewing angles to ensure comprehensive surface coverage. A minimum overlap of 60% between consecutive images was maintained to facilitate robust feature extraction and tie-point matching during photogrammetric reconstruction. 61 images were collected for the survey, each with a resolution of 6720 × 4480 pixels, to generate 3D models.

**Overall setup**: A tripod was used to stabilise the camera and maintain a fixed imaging height and angle. Physical calibration targets **(Figure 5)** were positioned around the specimen to provide a known scale reference. The distance between the object and the camera was fixed at different aspects, and artificial lighting was utilised to maintain consistent luminance, preserving the clarity and colour fidelity of the surface. These procedures established a robust workflow for high-quality imaging under laboratory conditions, ensuring that the generated model is consistent in both geometric and optical aspects, thereby supporting subsequent reconstruction and defect analysis.

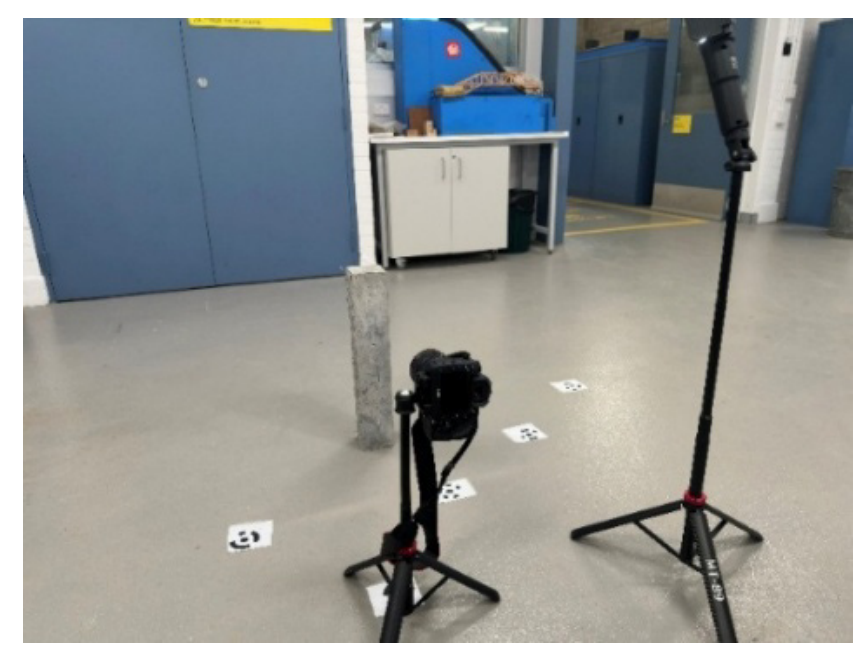

Figure 3 Data Acquisition Setup

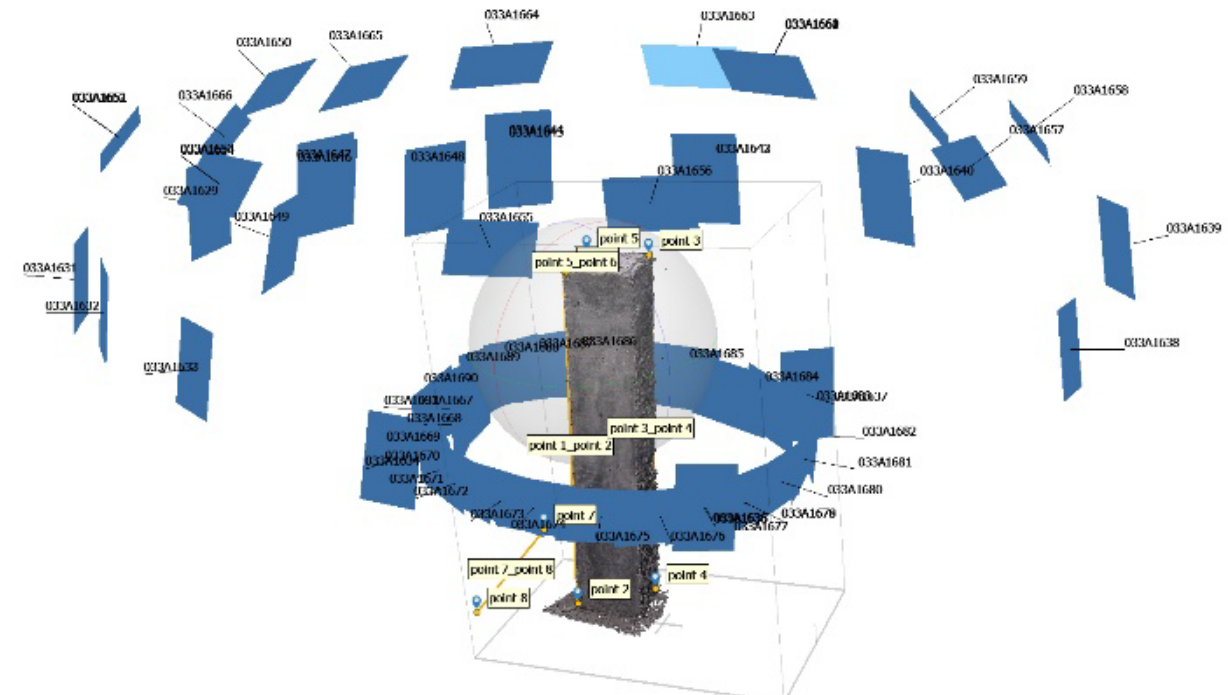

Figure 4 Capture Strategy

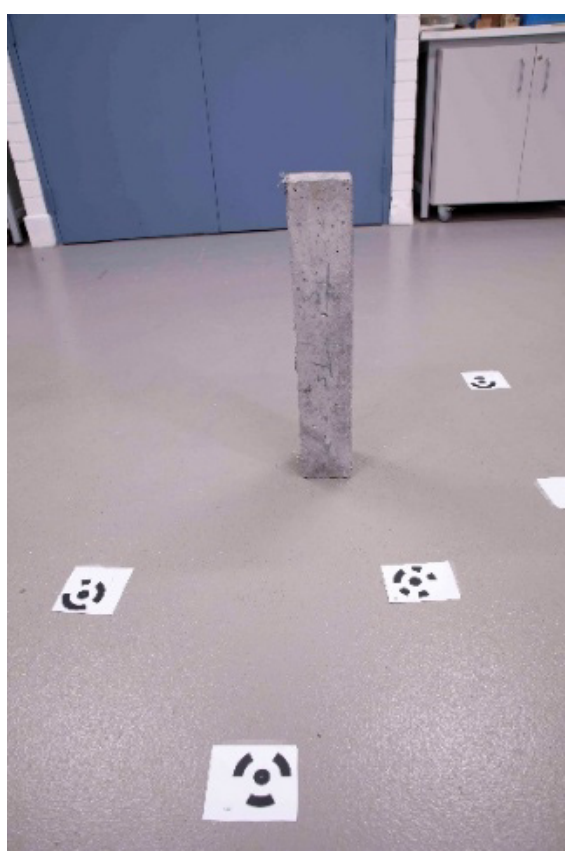

Figure 5 Scale Reference

### 3.3 3D reconstruction and analysis

A 3D point cloud reconstruction software, Agisoft Metashape Professional Edition (version 2.2.2), was employed to generate high-quality, well-textured, and detailed 3D data from the captured image set. The overall reconstruction workflow includes the following major stages: camera calibration, image alignment, scaling and optimisation, and dense point cloud generation.

***Camera calibration:*** Before image matching, the camera was calibrated to determine the intrinsic camera parameters, including focal lengths, principal point coordinates, and radial and tangential distortion coefficients. A checkerboard calibration target was used to correct lens distortion and improve geometric accuracy. To validate the automatic processing in Metashape, a similar procedure using MATLAB is conducted for assessment. The projective relationship between a 3D world coordinate point $(X, Y, Z)$ and its 2D image coordinate $(x, y)$ can be expressed as follows:

$$s\begin{bmatrix} u \\ v \\ 1 \end{bmatrix} = \begin{bmatrix} f_x & 0 & c_x \\ 0 & f_y & c_y \\ 0 & 0 & 1 \end{bmatrix} [RT] \begin{bmatrix} X_w \\ Y_w \\ Z_w \\ 1 \end{bmatrix}, \quad (1)$$

Where $s$ is a scaling factor, $R$ represents the radial coefficients, and $T$ represents the Tangential distortion.

***Image alignment:*** After calibration, SfM (also known as image alignment) was performed to identify and match key points across overlapping images. This step identifies and matches key points across overlapping images to determine camera positions and orientations, then generates a sparse point cloud. The spatial relationships between these matched points are then optimised through bundle adjustment to minimise the overall reprojection error and refine the camera geometry.

***Scaling and optimisation:*** After the initial sparse reconstruction, the model was rescaled and optimised using the known distances between control scale bars placed around the RuC specimen during image acquisition (as shown in **Figure 5**). These reference markers provide a metric constraint, allowing the software to convert the relative 3D geometry into true physical dimensions. Additional optimisation was carried out to adjust internal camera parameters and minimise residual errors across all tie points, ensuring accurate scaling consistency throughout the model.

***Dense 3D model generation*:** Once image alignment and scaling were complete, a dense reconstruction was carried out to generate a detailed 3D model. To address the inherent risk of artificial data interpolation, the point cloud densification parameters were strictly bounded by the physical optical resolution, specifically the GSD of 0.15 $mm$. Unlike pure mathematical interpolation, which simply estimates empty space between sparse laser measurements, the MVS algorithm derives 3D coordinates exclusively through the spatial intersection of homologous optical rays matched across multiple overlapping high-resolution images. By restricting the point generation to the validated optical limit rather than artificially oversampling between pixels, the resulting dense point cloud represents true physical surface textures rather than algorithmically generated noise. This strict optical constraint ensures that the extracted sub-millimetre crack geometries remain physically meaningful and metrically reliable.

## 4. Experiment results and discussions

### 4.1 3D model quality evaluation

To check the geometric precision and scaling reliability of the reconstructed 3D model, a comparison was conducted between the generated model and the real target. As shown in **Figure 6**, the width of a section of the crack was measured. The reconstructed crack width was measured at approximately 2.3 mm, aligning closely with the corresponding physical measurement of 2.36 $mm$. The accuracy of the calliper used in this paper is 0.02 $mm$, whereas manual measurements in dense point clouds are less accurate due to user interpretation and data noise. Nevertheless, the comparison results indicate that the photogrammetric workflow and scale calibration used in this study effectively reveal the target and provide high-resolution geometric and textural detail. This proves the reliability of utilising the reconstructed 3D model for subsequent defect analysis. To investigate the impact of camera resolution on point cloud reconstruction, an iPhone 16 and a Canon 5D Mark IV were used to capture images following the same workflow.

After aligning the models using the Iterative Closest Point (ICP) algorithm, the comparison indicates an average Cloud to Cloud (C2C) deviation of 0.82 $mm$ between the iPhone-generated model and the Canon DSLR reference model. Rather than pure geometric distortion, this deviation largely reflects the difference in hardware sensor noise and the high surface tortuosity of the RuC material. As shown in **Figure 7**, approximately 67.5% of the points deviate by less than 1 $mm$ compared with the DSLR-based model. Most discrepancies are concentrated around the edges or corners of the model, where key-point matching and image distortion may affect reconstruction accuracy. Given that the monitoring target in this study is millimetre-scale surface defects, the model generated from the DSLR camera is preferred.

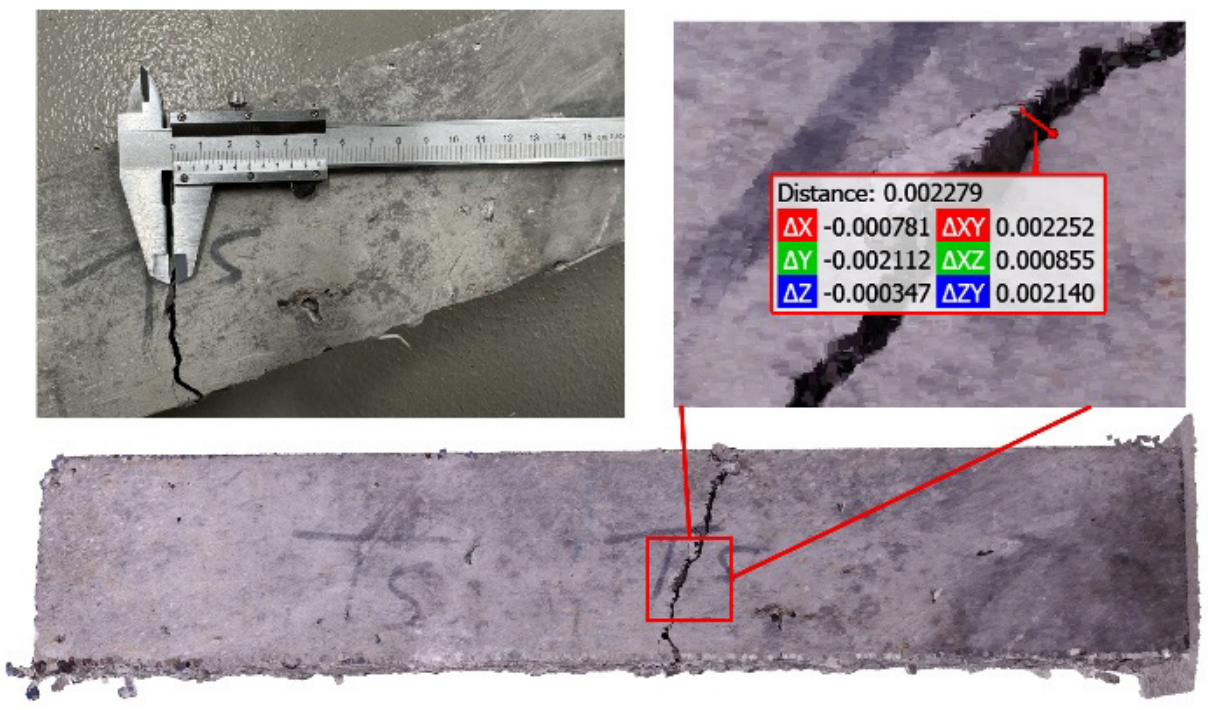

Figure 6 Width measurement

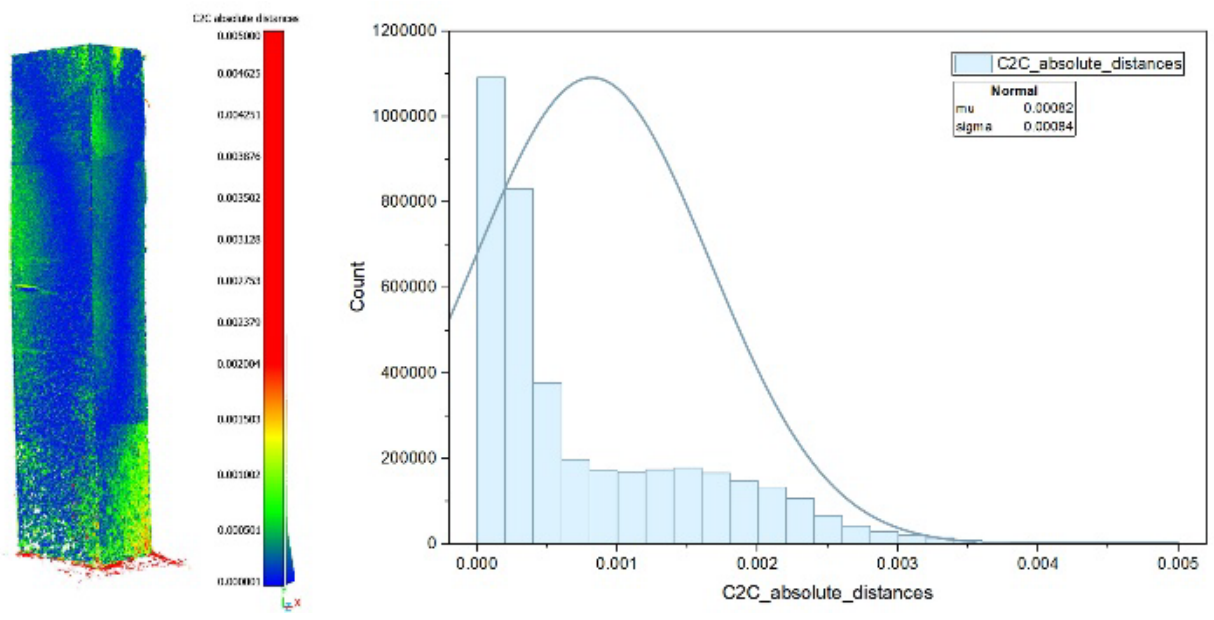

Figure 7 The Distance comparison between iPhone and Canon

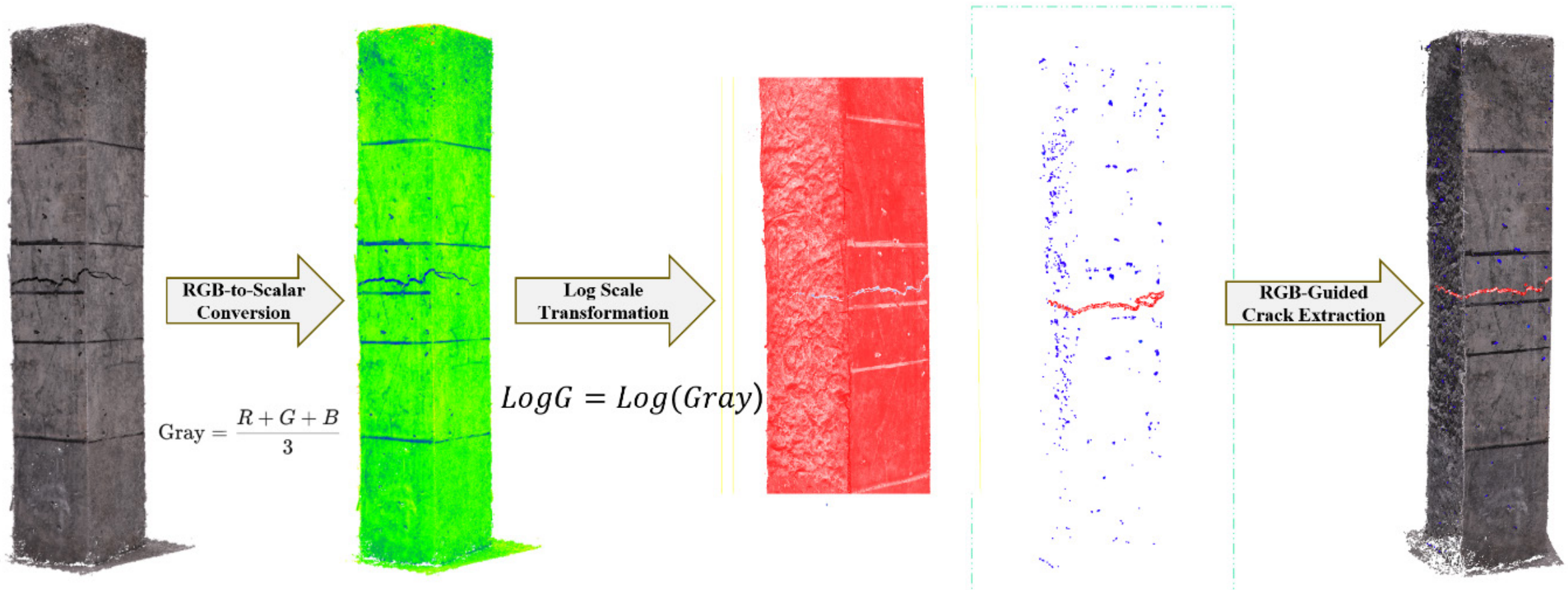

Figure 8 RGB-guided crack extraction

In terms of resolution, the average point spacing of the reconstructed models is 0.23 $mm$ for the iPhone and 0.05 $mm$ for the Canon DSLR. Although this spacing indicates sub-pixel algorithmic interpolation during the dense reconstruction phase, the effective optical resolution remains strictly bounded by the GSD of 0.15 $mm$. This highlighted how camera hardware performance, particularly physical sensor size and resolution, fundamentally governs the geometric quality of the 3D model. Therefore, a high-resolution DSLR camera equipped with a large CMOS sensor is highly recommended for applications requiring quantitative, fine-scale defect detection.

### 4.2 Crack identification

To isolate surface cracks from the background, an RGB-guided crack extraction method was employed. In this approach, the RGB colour information of the raw point cloud was first converted to grayscale by averaging the individual colour components. A logarithmic contrast enhancement was subsequently applied to amplify the subtle tonal differences between the intact surface and the darker crack areas, effectively maximising the distinction between the defect and normal points. While this enhancement highlights the visual boundaries of the crack, relying solely on a 2D planar mask is inadequate for capturing the surface variations caused by rubber pullout in RuC. Therefore, following this enhancement, an OPTICS clustering algorithm was implemented in 3D to group the data and isolate the true volumetric fracture points from residual background noise.

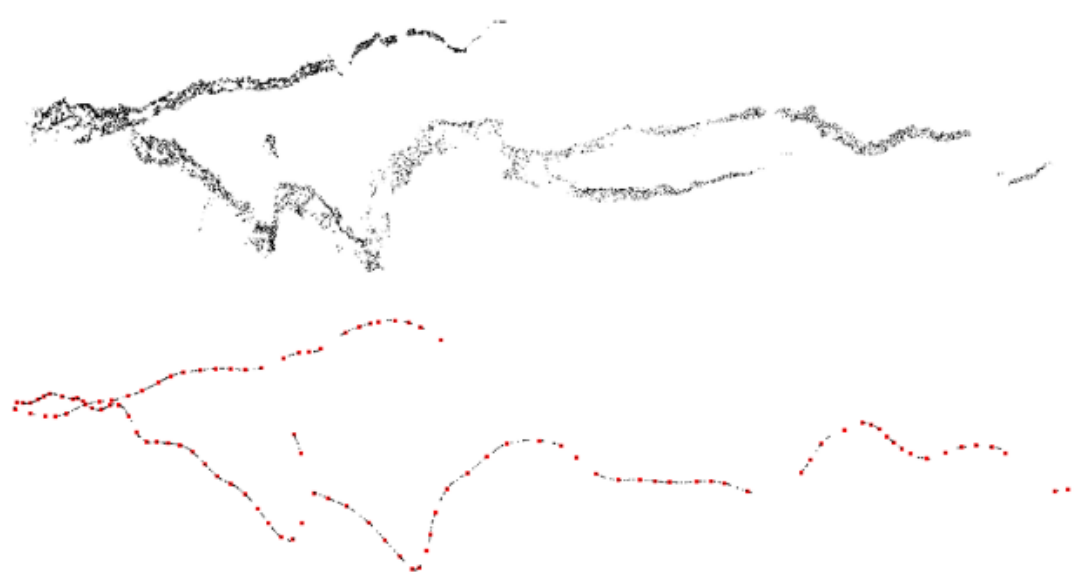

Figure 9 Extracted cracking points set

The processed grayscale data exhibit a distinct radiometric discrepancy between the true crack points and the surrounding background (**Figure 8**). This contrast enables the spatial isolation of the fracture point cloud, preparing the dataset for subsequent geometric analysis. The extracted crack points maintain a well-structured geometric representation (**Figure 9**), providing a high-fidelity point set which is suitable for precise morphological interpretation. To facilitate quantification of the 3D dimension, a skeleton-based algorithm will subsequently be employed to simplify the complex geometry of these extracted points into a measurable spatial network.

### 4.3 Pre- and Post-Test deformation analysis

Finally, a pre-test and post-test deformation analysis was conducted to quantify the surface displacement caused by the four-point bending test. To achieve this, two sets of image sequences were captured: one pre-test and one post-test. Following the workflow described in Section 3, two 3D models were generated.

By computing the Cloud-to-Cloud (C2C) distances between the pre-test and post-test models, the magnitude of the surface deviation can be visually and metrically quantified (**Figure 10**). The spatial colour mapping indicates that the maximum displacement occurs in the primary fracture zone, directly corresponding to the region of visible cracking. Within this area, the average displacement was approximately 2.2 $mm$, with the majority of deviation values ranging from 1.3 $mm$ to 2.9 $mm$

These results demonstrate that the photogrammetric workflow not only identifies defect points but also provides quantitative, spatially continuous measurements of surface deformation. This approach highlights the potential of close-range 3D reconstruction as a non-contact, flexible, comprehensive, and accurate method for structural monitoring and damage assessment in laboratory environments.

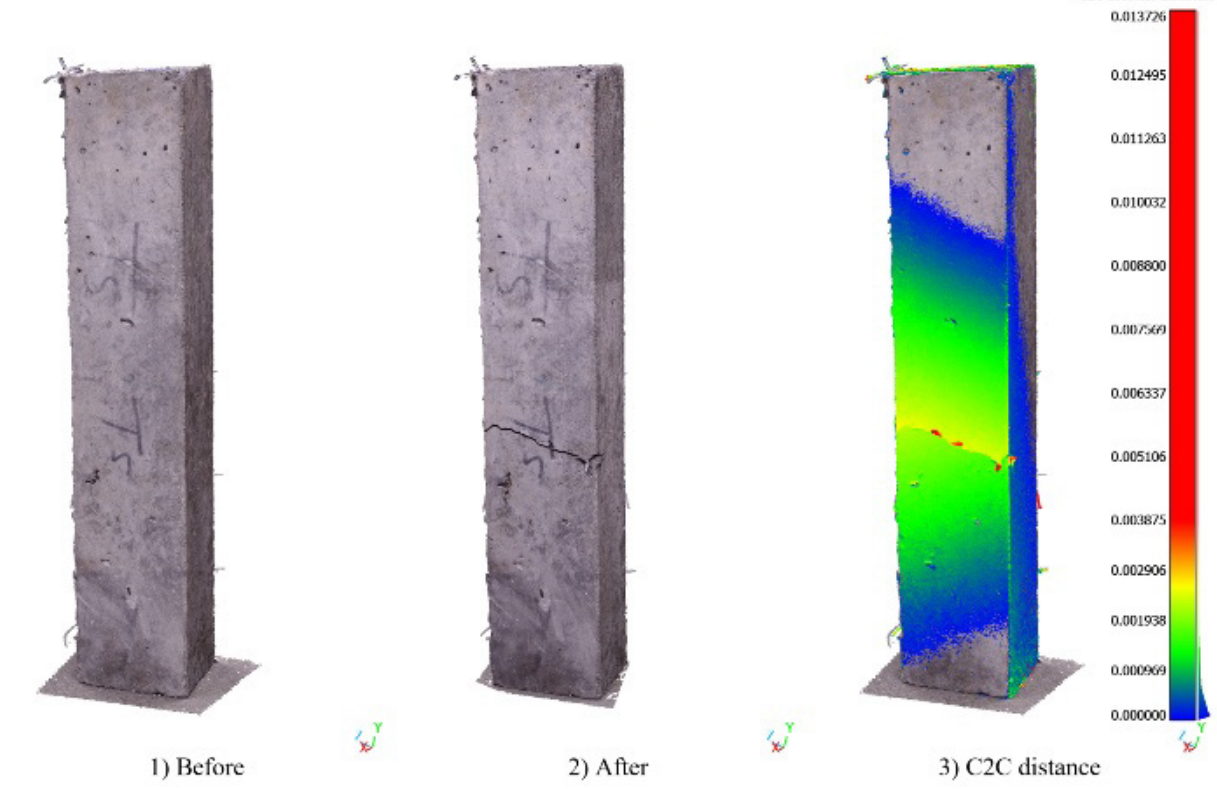

Figure 10 Deformation comparison

## 5. Conclusion and future work

Nowadays, 3D point cloud data has become a common dataset in civil engineering, providing comprehensive geometric information and enhanced visualisation for both academic and industrial applications. This paper demonstrates the feasibility and performance of close-range photogrammetric approaches for analysing minor defects and deformation in RuC samples. 3D reconstruction software based on SfM technology enables the generation of dense point clouds with sub-millimetre resolution, providing an alternative to TLS or MLS for fine-scale defect monitoring. Specifically, this paper first compares reconstruction results with real physical measurements across different imaging equipment, confirming that high-resolution cameras with higher pixel counts yield more detailed models. Subsequently, cracked regions were segmented from the point cloud for dimensional estimation, and millimetre-scale surface displacements were quantified through a pre- and post-test deformation analysis. The results confirm that the photogrammetric method can be considered as an accessible and suitable tool for structural analysis, particularly at the micro-defect level. By overcoming the planar limitations of traditional 2D imaging, this 3D workflow successfully captures the highly tortuous fracture morphologies unique to RuC, offering an approach to better understand, analyse, and ultimately optimise sustainable construction materials.

For future work, three primary research directions can be developed. **First**, state-of-the-art (SOTA) learning-based reconstruction algorithms, such as Neural Radiance Fields (NeRF) and 3D Gaussian Splatting, can be investigated and compared with traditional SfM methods to explore more advanced and efficient data-generation solutions. Because these novel view synthesis techniques inherently face challenges regarding absolute metric scale and physical validation, future methodologies will explore integrating the current workflow with handheld LiDAR scanners to establish a rigorous geometric baseline. **Second**, because manual feature extraction remains labour-intensive and subject to micro-scale noise, an automated, algorithmic 3D feature extraction pipeline will be developed. This pipeline will systematically estimate all spatial dimensions of the detected defects using a skeleton-based representation, including length, width, and depth, to fully exploit the point cloud data and create a comprehensive digital twin of each specimen. **Finally**, dynamic crack propagation over time monitoring can be achieved by building a multi-camera system or employing videogrammetry-based techniques.

### Acknowledgement

This research was partly supported by the China Scholarship Council (CSC).